\documentclass[numbers]{article}

\usepackage{float}
\usepackage[preprint]{neurips_2025}

\usepackage[utf8]{inputenc} 
\usepackage[T1]{fontenc}    
\usepackage{hyperref}       
\usepackage{url}            
\usepackage{booktabs}       
\usepackage{amsfonts}       
\usepackage{nicefrac}       
\usepackage{microtype}      
\usepackage{xcolor}         
\usepackage{amsmath}
\usepackage{enumitem}
\usepackage{subcaption}
\usepackage{graphicx, subcaption}
\usepackage{makecell}       
\usepackage{booktabs}   
\usepackage{array}     
\usepackage{makecell}  
\usepackage{multirow}   
\usepackage{lineno}
\usepackage{algorithm}     
\usepackage{algpseudocode}   
\algrenewcommand\algorithmiccomment[1]{\hfill$\triangleright$~#1}  

\title{FHGS: Feature-Homogenized Gaussian Splatting}

\author{%
  Q. G. Duan\  \
  Benyun Zhao\ \ 
  Mingqiao Han\ \ 
  Yijun Huang \ \
  Ben M. Chen\textsuperscript{*}   \\
  Department of Mechanical and Automation Engineering \\
  The Chinese University of Hong Kong \\
  \texttt{\{qigenduan, byzhao, mqhan, yjhuang, bmchen\}@cuhk.edu.hk} \\
  \begin{small}
    \textsuperscript{*}Corresponding Author
  \end{small}
}

\begin{document}

\maketitle

\begin{abstract}
Scene understanding based on 3D Gaussian Splatting (3DGS) has recently achieved notable advances. Although 3DGS related methods have efficient rendering capabilities, they fail to address the inherent contradiction between the anisotropic color representation of gaussian primitives and the isotropic requirements of semantic features, leading to insufficient cross-view feature consistency.
To overcome the limitation, we proposes \textit{FHGS} (Feature-Homogenized Gaussian Splatting), a novel 3D feature fusion framework inspired by physical models, which can achieve high-precision mapping of arbitrary 2D features from pre-trained models to 3D scenes while preserving the real-time rendering efficiency of 3DGS.
Specifically, our \textit{FHGS} introduces the following innovations: Firstly, a universal feature fusion architecture is proposed, enabling robust embedding of large-scale pre-trained models' semantic features (e.g., SAM, CLIP) into sparse 3D structures.
Secondly, a non-differentiable feature fusion mechanism is introduced, which enables semantic features to exhibit viewpoint independent isotropic distributions. This fundamentally balances the anisotropic rendering of gaussian primitives and the isotropic expression of features; Thirdly, a dual-driven optimization strategy inspired by electric potential fields is proposed, which combines external supervision from semantic feature fields with internal primitive clustering guidance. This mechanism enables synergistic optimization of global semantic alignment and local structural consistency.
Extensive comparison experiments with other state-of-the-art methods on benchmark datasets demonstrate that our \textit{FHGS} exhibits superior reconstruction performance in feature fusion, noise suppression, and geometric precision. 
More interactive results can be accessed on:~\url{https://fhgs.cuastro.org/}. 
\end{abstract}

\begin{figure}[htbp]
  \centering
  \includegraphics[width=1.0\textwidth]{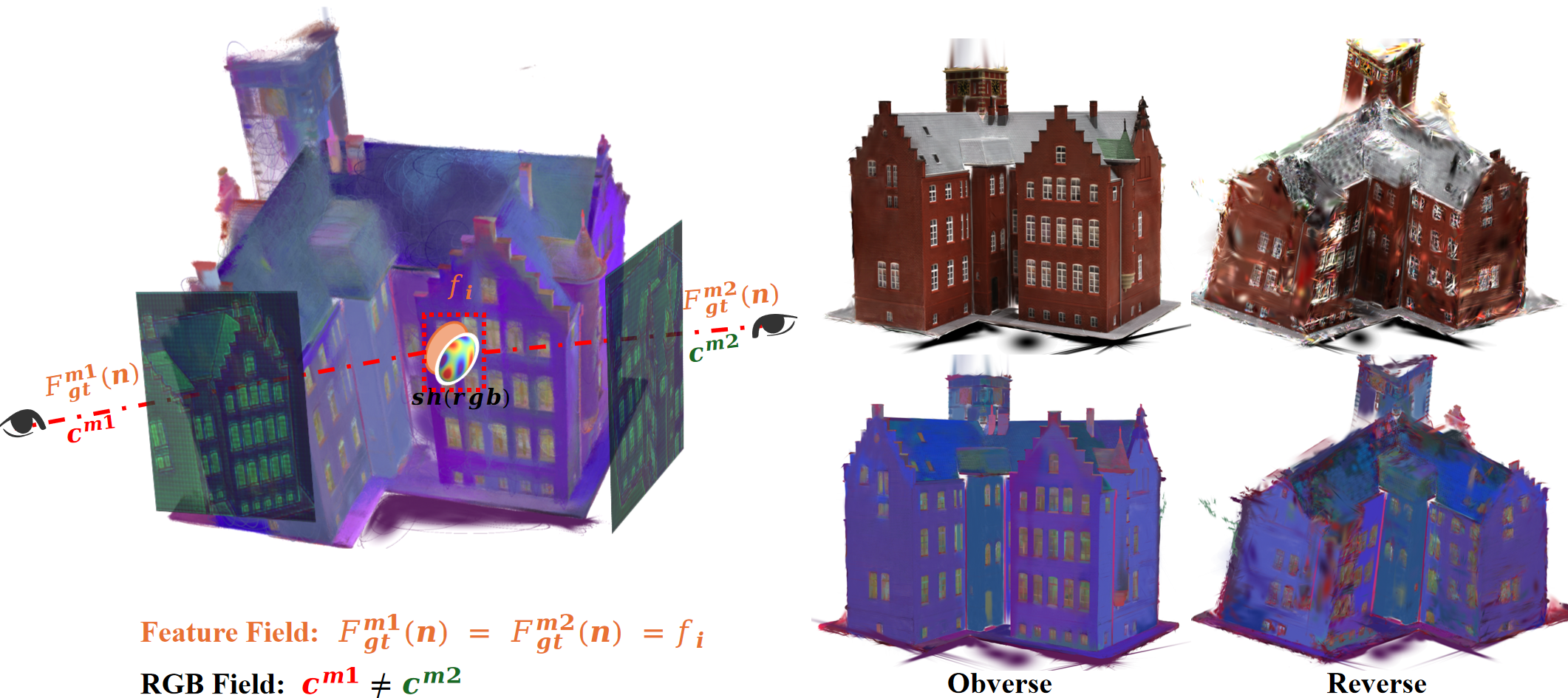} 
  \caption{The left part demonstrates the inherent contradiction between the anisotropic color $\mathbf{c}$ of gaussian primitives in RGB field and the isotropic requirement of semantic features $\mathbf{f}$. The right part shows the results in obverse and reverse, indicating that \textit{FHGS} shows superior reconstruction performance in terms of feature fusion, noise suppression, and geometric accuracy.}
  \label{fig:global}
\end{figure}

\section{Introduction}

In recent years, scene representation—particularly understanding—has emerged as a prominent research focus, as it enables unmanned systems to better perceive and interpret their surrounding environments. Traditional scene representation frameworks such as Multi-View Stereo~\cite{MVSNet} (MVS) and Simultaneous Localization and Mapping~\cite{fastlivo} (SLAM) can achieve geometric reconstruction. However, these methods rely on non-differentiable pipelines and remain limited in high-level semantic perception and nonlinear feature fusion. As a result, the differentiable approaches have gradually come into focus. Among them, Neural Radiance Fields (NeRF)~\cite{nerf} and 3D Gaussian Splatting (3DGS)~\cite{3Dgaussians} have revolutionized the scene representation framework. NeRF models implicit radiance fields and learns continuous 3D spatial representations under 2D image supervision via differentiable volume rendering equations, whereas 3DGS adopts explicit anisotropic Gaussian primitives to enable high-quality reconstruction through efficient rasterization. 
However, these traditional neural field representations primarily focus on the fusion of RGB geometric fields, with limited exploitation of semantic features. In contrast, feature fields require maintaining semantic consistency across multiple viewpoints to prevent contradictory predictions during viewpoint transitions \cite{tvcg}, as shown in Fig.~\ref{fig:global}.

Against this backdrop, the integration of semantic feature from transformer-based models~\cite{vit,mae} fusing with NeRF and 3DGS framework has begun to emerge. NeRF-based frameworks extend radiance fields by incorporating learnable semantic feature fields, implicitly enforcing multi-view semantic consistency through continuous neural representations~\cite{panoptic_lift, zhi2021place}. However, their inference speed remains limited due to the dense sampling required by volumetric rendering. In contrast, 3DGS-based frameworks~\cite{featuregs} construct explicit feature fields by directly associating semantic features with their corresponding explicit primitives. However, as shown in the Fig.~\ref{fig:global}, an inherent conflict arises between the anisotropic nature of RGB fields in their rasterization pipeline and the isotropic representation required for robust semantic features. Existing methods, whether based on implicit or explicit representations, typically treat features as fully differentiable and optimize them jointly with appearance. However, this continuous optimization may introduce inconsistencies that interfere with the self-attention mechanism in transformer, leading to feature noise and degraded rendering quality.

To address the limitations of aforementioned phenomena, we propose \textit{FHGS} (Feature-Homogenized Gaussian Splatting), a novel feature fusion framework built upon the GS paradigm which establishes bidirectional associations between 2D semantic features and 3D feature fields, enabling end-to-end optimization for multi-view consistent feature fusion. \textit{FHGS} preserves the efficiency and explicitness of the gaussian splatting while overcoming the limitations of rasterization-based methods designed primarily for RGB reconstruction. Specifically, each gaussian primitive is augmented with non-differentiable semantic features, which are directly supervised by ground-truth feature maps to enforce semantic consistency across views. To better achieve multi-view feature consistency under efficient optimization while preserving isotropic representations within the feature field, we propose a dual-driven mechanism inspired by physics-inspired principles from electric field modeling. This mechanism, composed of \textit{External Potential Field Driving} and \textit{Internal Feature Clustering Driving}, constrains anisotropy to photometric properties merely, while enforcing isotropy in the feature field to support consistent semantic representation. 

Extensive experiments of benchmark datasets demonstrate the proposed \textit{FHGS} not only enhances semantic fusion quality but also improves geometric reconstruction accuracy and noise robustness through feature-driven regularization effects. The main contributions of this work are as follows:

\begin{itemize}
\item General feature fusion architecture: We propose a GS-based feature field fusion framework capable of integrating 2D semantic features extracted from large-scale pre-trained models (e.g., SAM~\cite{sam}, CLIP~\cite{clip}), enabling unified optimization from low-level geometry to high-level semantics.
\item Integration of non-differentiable features into GS framework: We pioneer about integrating of non-differentiable features into the differentiable gaussian splatting methods, which fundamentally resolves the inherent contradiction between the anisotropic nature of gaussian primitives and the isotropic requirements of semantic features.
\item Physics-inspired dual-drive mechanism: Inspired from electric field modeling, we design a joint optimization strategy combining external potential field driving and internal feature clustering driving, characterized by intuitive logic, computational efficiency, and strong interpretability. Additionally, the metric based on this mechanism, named FE, is proposed to evaluate the global consistency of features.
\item Performance superiority: Compared with other 3DGS feature fusion frameworks on benchmark datasets, our \textit{FHGS} achieves state-of-the-art fusion performance, and optimizes the performance of geometric reconstruction.
\end{itemize}

\section{Related Work}
\label{sec:related}
\subsection{Novel View Synthesis}

Neural Radiance Fields (NeRF)~\cite{nerf} models a continuous 3D scene representation through an implicit radiance field and a differentiable volume rendering equation supervised by 2D images. The core of NeRF is that it leverages a multilayer perceptron (MLP) to map spatial positions and viewing directions to color and density values, enabling novel view synthesis with high-quality via ray integration. Subsequent works such as Mip-NeRF~\cite{mipnerf}, Instant-NGP~\cite{instantngp}, and Mip-NeRF 360~\cite{mipnerf360} further improve anti-aliasing, training speed, and scalability to large-scale unbounded scenes. 
However, the nature of implicit representation of NeRF-based methods~\cite{mipnerf, instantngp, mipnerf360} requires dense sampling and complex network inference, leading to low training and rendering efficiency that limits its applicability in real-time scenarios.

To overcome the limitations of implicit representations, 3D Gaussian Splatting (3DGS)~\cite{3Dgaussians} introduces an explicit scene representation by decomposing the 3D environment into a set of explicit, anisotropic gaussian primitives.
Combined with the implicit pipeline, this formulation enables efficient training and rendering. Moreover, the anisotropic view-dependent appearance is further represented using spherical harmonics (SH), conditioned on the spatial and radiometric properties of each primitive. Compared to NeRF, 3DGS eliminates the need for neural networks in the rendering pipeline, significantly improving memory efficiency and real-time performance while maintaining high-fidelity reconstruction. Extending this idea, 2D Gaussian Splatting (2DGS)~\cite{2dgs} enhances multi-view geometric consistency by anchoring gaussian primitives to the image plane while enforcing depth consistency constraints. Despite these advances, existing GS-based frameworks remain primarily focused on geometry reconstruction, without addressing a core challenge of the utilization of semantic features.
Our method \textit{FHGS} introduces high-level semantic priors from SAM~\cite{sam}, enabling structure-aware guidance during reconstruction.
Different from conventional GS methods that rely purely on differentiable photometric cues, our \textit{FHGS} leverages non-differentiable, high-dimensional semantic information to guide the optimization of semantic-aware structural distributions, resulting in more precise and robust reconstructions, especially in challenging regions where appearance cues alone are insufficient.

\subsection{Implicit Feature Fusion}
Integrating semantic information or learned features into point-based scene representations is a well-established strategy, extensively explored in the NeRF-based works~\cite{nerfsos,lerf,Dnerf,3dv,Inplace,P3su} and now migrating to gaussian splatting. Recent attempts to incorporate features into 2D/3D gaussian splatting can be clustered into three categories. 
Mask fusion approaches exemplified by GaussianCut~\cite{gaussiancut} and Gaussian Grouping~\cite{gaussian_grouping} provided 2D masks onto the gaussian primitives set and employ graph-cut or low-dimensional identity embeddings for partitioning. While this method is straightforward for interactive 2D editing, the resulting correspondence is no longer perceptually obvious in 3D space, and these method still depends on extensive manual annotation, which fails to capture any high-dimensional semantic features. 
External fusion schemes such as SAGA~\cite{saga}, Semantic Gaussians~\cite{semanticgaussian} and OmniSeg3D~\cite{omniseg3d} distilled 2D features into 3D space through an auxiliary neural network or contrastive learning, thereby enriching semantic information at the expense of additional parameters, prolonged training, and deviation from the concise design philosophy of gaussian splatting. 
Feature fusion techniques including Feature 3DGS~\cite{featuregs} and LangSplat~\cite{langsplat} learned embeddings with individual primitive so that semantics render with color. However, these embeddings overwrite or reshape the high-dimensional tensors supplied by large segmentation models, erasing the self-attention structure and class relationships encoded therein and often introducing substantial noise that degrades segmentation quality. As a consequence, subsequent reasoning is confined to image space rather than the gaussian primitives domain. 

Therefore, we model cross-view semantic coherence as a physics-inspired potential-field optimization that relocates gaussian primitives while preserving their original feature vectors. The pipeline of our \textit{FHGS} is self-supervised and globally consistent, retains the full high-dimensional semantic tensor for downstream tasks such as segmentation, detection and multi-modal prompting, and maintains the real-time rendering performance fundamental to gaussian splatting.

\section{Methodology}

The proposed \textit{FHGS} addresses the semantic distortion and efficiency bottlenecks caused by the conflict between the anisotropic rendering mechanism of gaussian splatting and the isotropic requirements of high-level semantic features. There are three core components of \textit{FHGS}, which will be successively illustrated  in this section:
(1) A general-purpose feature fusion architecture that supports the integration of multi-view features.
(2) A GS framework enhanced with non-differentiable features, enabling the incorporation of high-dimensional semantic priors.
(3) A dual-driven feature fusion mechanism inspired by physical modeling, which guides the feature optimization process using both geometric and semantic consistency cues.

\subsection{General Feature Fusion Architecture}
\begin{figure}[htbp]
  \centering
  \includegraphics[width=0.9\textwidth]{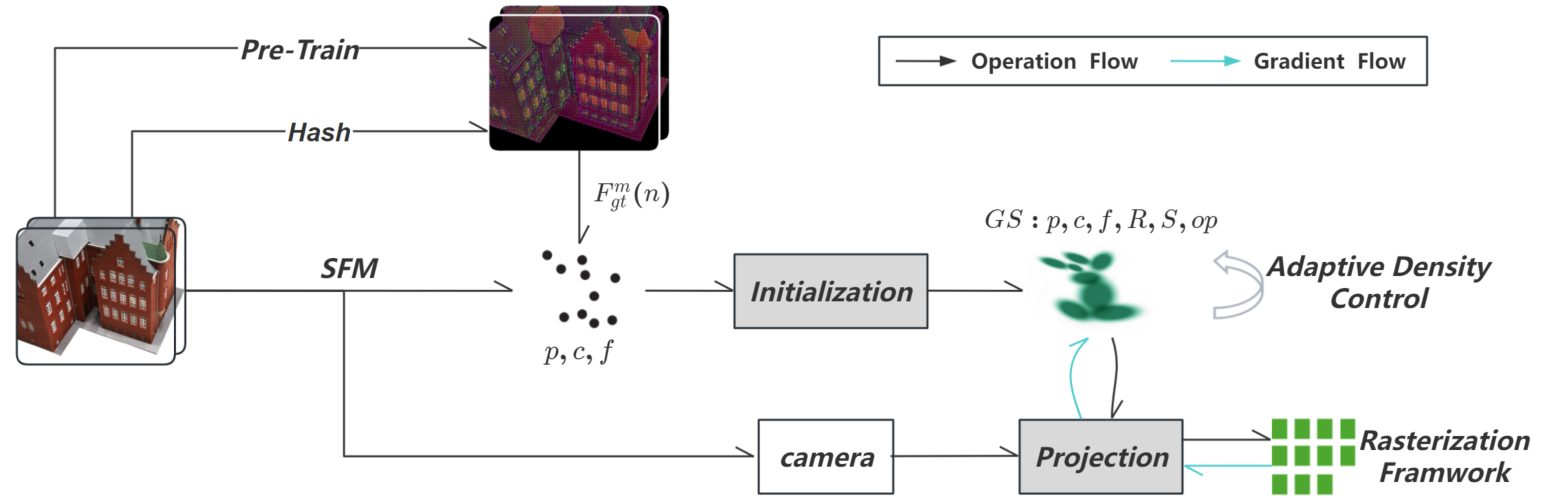} 
  \caption{Pipeline of the General Feature Fusion Architecture}
  \label{fig:method_f1}
\end{figure}

The pipeline of the General Feature Fusion Architecture is demonstrated in Fig.~\ref{fig:method_f1}: The Structure from Motion (SFM) process reconstructs a sparse 3D point cloud \(PC\) via Bundle Adjustment (BA) firstly. To accelerate the correspondence of 3D point cloud and 2D feature, we construct a spatial hash table \(\mathcal{H}\) that indexes the projections of each 3D point \(pc_i\) across visible views \(M\).
Subsequently, pre-trained model of segmentation is used to generate 2D ground-truth feature maps \(\mathbf{F}_{gt}^m\). Given a 3D point \(pc_i\), its corresponding pixel \(n\) in a randomly selected view $m \in M$ is retrieved via the spatial hash table, and the semantic feature \(\mathbf{f}_i = \mathbf{F}_{gt}^m(n)\) is sampled accordingly.  
Each gaussian primitive is initialized from a point \(pc_i\) and inherits its geometric parameters, appearance attributes, and a frozen semantic feature. Specifically, a primitive is anchored at a center position \(\mathbf{p}_i\) and oriented by two orthogonal tangent directions \(\mathbf{t}_u\) and \(\mathbf{t}_v\), with the normal vector defined as \(\mathbf{t}_w = \mathbf{t}_u \times \mathbf{t}_v\). These directions form the rotation matrix \(\mathbf{R}_i = [\mathbf{t}_u, \mathbf{t}_v, \mathbf{t}_w] \in \mathbb{R}^{3 \times 3}\). The spatial extent on the tangent plane is described by a planar scale vector \(\mathbf{S}_i = (s_u, s_v)\). For appearance modeling, each primitive carries an RGB color \(\mathbf{c}_i \in \mathbb{R}^3\) and an opacity scalar \(op_i \in [0, 1]\). Additionally, a frozen semantic feature vector \(\mathbf{f}_i\) is assigned to each primitive, extracted via a non-differentiable image embedding. We write the complete representation of each gaussian primitive as:
\begin{equation}
    \theta_i = \{\,\mathbf{p}_i, \mathbf{R}_i, \mathbf{S}_i, op_i, \mathbf{c}_i, \mathbf{f}_i\,\}.
\end{equation}
The other symbols follow the notation of conventional 3DGS. Any point \((u,v)\) in the tangent plane is mapped to world space by:  
\begin{equation}
        P(u, v) = \mathbf{p}_i + s_u \mathbf{t}_u u + s_v \mathbf{t}_v v = \mathbf{H}(u, v, 1, 1)^\top
\end{equation}
with the homogeneous matrix \( \mathbf{H} \in \mathbb{R}^{4 \times 4} \) factories translation, rotation and scale. Given local coordinates \(\mathbf u=(u,v)\), the unnormalized density is $\mathcal{G}(\mathbf{u}) = \exp(-\frac{u^2+v^2}{2})$. Then, let \(\mathbf{x}=(x,y)\) be a pixel and define \(\mathbf{u(x)}\) as the unique point in the splat's tangent plane whose homogeneous coordinates satisfy:
\begin{equation}
    \mathbf x=(xz,\;yz,\;z,\;1)^{\!\top}=\mathbf{W}P(u,v)=\mathbf{W}\mathbf{H}(u,v,1,1)^{\!\top}
    \label{equ:x}
\end{equation}
where \(\mathbf{W} \in \mathbb{R}^{4 \times 4}\) is the world-to-camera transformation matrix, and \(z\) denotes the depth. During the rasterization process in the GS framework, primitives that intersect with the ray \(l\) emitted from pixel \(n\) are identified. Specifically, the \(N\) primitives covered by ray \(l\) are sorted by their rendering depth, with index \(i=1\) and \(i=N\) assigned to the farthest and the nearest, respectively. The final color can be computed as:
\begin{equation}
    \mathbf{c}(\mathbf{x}) 
    = \sum_{i=1}^{N} \mathbf{c}_i\ w_i
\end{equation}
The weight \({w}_i = \alpha_i T_i\) is the dynamic differentiable parameter, while \(\alpha_i = op_i{\mathcal{G}}_i(\mathbf{u}(\mathbf{x}))\) characterizes the intrinsic properties of gaussian primitives, and  \(T_i = \prod_{j=1}^{i-1}(1-\alpha_j)\) encodes their transmittance. During the backward propagation, gradients of \(w_i\) propagate through the chain rule to drive the optimization of the geometric parameters of gaussian primitives, thereby enhancing scene reconstruction quality. As the pivotal variable linking geometry and the differentiable rasterization, \(w_i\) directly drives both reconstruction accuracy and rendering efficiency.

\subsection{Non-Differentiable Features Fusion Mechanism}
\begin{figure}[htbp]
  \centering
  \includegraphics[width=0.9\textwidth]{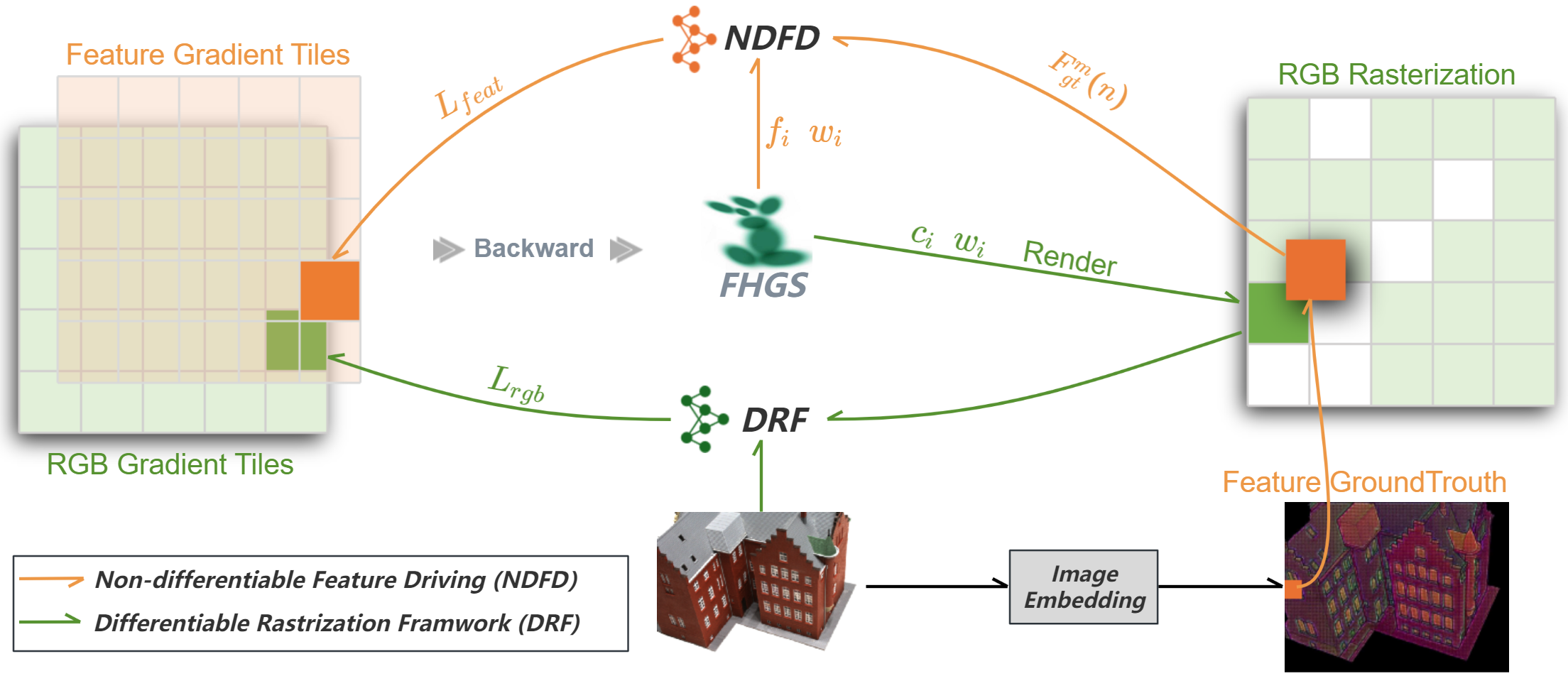} 
  \caption{Schematic representation of the two mechanisms of \textit{FHGS}: NDFD and DRF}
  \label{fig:rasterization}
\end{figure}

\textit{FHGS} integrates a non-differentiable feature driving (NDFD) (\textcolor{orange}{orange} arc pathway in Fig.~\ref{fig:rasterization}) with the original GS framework. During the forward process, \textit{FHGS} directly utilizes \(\mathbf{F}_{gt}^m\) compute the feature loss \(L_{feat}\) based on the feature \(\mathbf{f}_{i}\) and contribution weights \(w_{i}\). It is worth noting that the forward process does not require prior feature rendering, which can further reduce the computational costs. In the backward process, although the feature \(\mathbf{f}_{i}\) of each gaussian primitive is non-differentiable, the gradient of \(L_{feat}\) can still propagate through \(w_i\) to optimize \(\{\,\mathbf{p}_i, \mathbf{R}_i, \mathbf{S}_i, op_i\,\}\), implicitly guiding gaussian primitives toward feature-consistent regions.

Compared to the differentiable rasterization framework (DRF) in the conventional GS methods (\textcolor{green}{green} arc pathway in Fig.~\ref{fig:rasterization}), the non-differentiable branch eliminates the need for feature rendering during the forward process, enabling direct loss computation while preserving the efficiency of GS framework. 
This design brings the following characteristics: the anisotropic color rendering remains dedicated to illumination and shadow modeling, while the multi-view consistency of non-differentiable features is achieved through \(w\)-driven distribution optimization, thereby avoiding direct conflicts between the rasterization anisotropic mechanism and the isotropic requirements of semantic features. The detailed pseudo code of NDFD is given in the Appendix.

\subsection{Physics-Inspired Dual-Drive Mechanism}
\begin{figure}[htbp]
  \centering
  \includegraphics[width=0.8\textwidth]{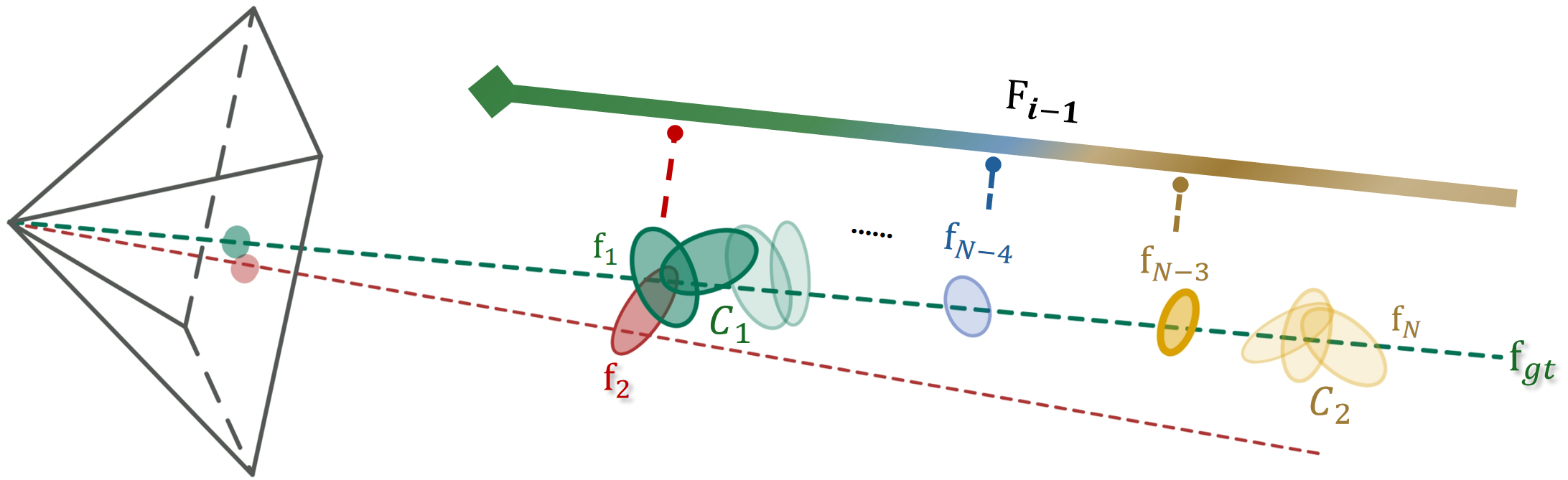} 
  \caption{The illustration of proposed Dual-Drive Mechanism: The color of each gaussian primitive and ray represents their feature properties, and the transparency represents the magnitude of the weight $w_i$ of the primitive on the ray. \(\mathbf{f}_{N-3}\) exhibits similarity to posterior accumulated values \(\mathbf{F}_{i-1}\), which is the value of the cluster $C_2$ only constrained by \(L_{gt}\); \(\mathbf{f}_{N-4}\) represents the inter-cluster noise points of $C_1$ and $C_2$ suppressed by \(L_{gt}\) and \(L_{cf}\); \(\mathbf{f}_2\) corresponds to the internal noise points from cluster $C_1$, where both \(L_{gt}\) and \(L_{cf}\) effectively optimize the distribution of $C_1$.}
  \label{fig:camera}
\end{figure}

Inspired by principles from an intuitive analogy in physical field theory, we model the \(\mathbf{F}_{gt}^m\) within the rasterization as a feature field in homogeneous space \(\mathbf{x}\), as defined in the Eq.~\ref{equ:x}, 
we consider it as an "electric field". More concretely, as illustrated in the Fig.~\ref{fig:camera}, we treat the ray \(l\) emitted from pixel \(n\) as an electric field line, and define its semantic property as the ground-truth feature \(\mathbf{f}_{gt} = \mathbf{F}_{gt}^m(n)\) sampled from the 2D feature map. The gaussian primitives are conceptualized as discrete "charges" carrying intrinsic features \(\mathbf{f}_i\). The feature loss  \(L_{feat}\) is then formulated as the potential energy loss in this electric field analogy. During the backward process, gradients drive the spatial optimization of gaussian primitives, analogous to the motion of charges under electric field forces toward regions of lower potential energy.

\emph{External Potential Field Constraint}: Following the logic of NDFD, we compute the cumulative similarity between the features  \(\mathbf{f}_i\) intersecting with ray  \(l\) and the ground-truth feature \(\mathbf{f}_{gt} \), constructing a similarity loss during the forward process: 
\begin{equation}
    L_{gt} = \sum_{i=1}^{N} w_i \sigma_i
    \label{equ:lgt}
\end{equation}
To eliminate the inherent contradiction between gaussian primitives and ground-truth semantics in the feature space, \textit{FHGS} introduces a similarity-based activation function \(\sigma_i = \frac{1}{1 + e^{k(\varphi - \lambda)}}\), where \(\varphi = \cos\left\langle \mathbf{f}_i, \mathbf{f}_{gt} \right\rangle\). This sigmoid function maps feature similarity into a polarity-like response, analogous to the binary behavior of electric charges. More detailed explanation of sigmoid function are given in the Appendix.

\emph{Internal Clustering Driving:} 
In order to suppress noise, enhance semantic coherence, and quantify the semantic feature entropy at pixel $n$, we simplify the bidirectional traversal of feature similarity between gaussian primitives during the rasterization process as:
\begin{equation}
\begin{split}
    L_{cf} 
    &= \sum_{i=1}^{N}\sum_{j=1}^{i-1}\sigma_i w_i w_j\left(1 - \mathbf{f}_i \cdot \mathbf{f}_j\right) \\
    &= \sum_{i=1}^{N}\sigma_i w_i\left(W_{i-1} - \mathbf{F}_{i-1} \cdot \mathbf{f}_i\right)
\end{split}
\end{equation}
The detailed derivation process of $L_{cf}$ can be found in the Appendix.
Since both \(\mathbf{f}_i\) and \(\mathbf{f}_{j}\) are normalized, \(\cos\left\langle \mathbf{f}_i, \mathbf{f}_{j} \right\rangle\>\)simplifies to \(\mathbf{f}_i\cdot \mathbf{f}_{j}\). We can obtain the cumulative weight \(W_n = \sum_{i=1}^{n} w_i\) and cumulative feature \(\mathbf{F}_n = \sum_{i=1}^{n} w_i \mathbf{f}_i\) along the ray from far to near. In addition, each current feature \(\mathbf{f}_i\) is compared only with the cumulative feature \(\mathbf{F}_{i-1}\). This avoids spurious repulsion across unrelated objects and reduces complexity from \(\mathcal{O}(N^2)\) to \(\mathcal{O}(N)\).  Furthermore, the cumulative weight \(W_{i-1}\) encodes the rendering contributions of farther gaussian primitives, implicitly modeling depth hierarchy. More specifically, the similarity activation function \(\sigma_i\) suppresses interference from background clutter, preventing incorrect contributions to foreground semantic clusters (e.g., $C_1$ and \(\mathbf{f}_1\) in the Fig.~\ref{fig:camera}). This mechanism achieves local semantic coherence by anchoring primitives that are semantically consistent with \(\mathbf{f}_{gt}\) (e.g., $\mathbf{f}_1$), while repelling dissimilar ones, thereby reducing feature conflicts and reinforcing cluster purity. It effectively suppresses internal noise (e.g., \(\mathbf{f}_2\) in the Fig.~\ref{fig:camera}) and eliminates irrelevant outliers in space (e.g., $C_2$ and \(\mathbf{f}_{N-3}\) in the Fig.~\ref{fig:camera}), resulting in cleaner and more compact semantic regions.

The aforementioned two driving methods, together with \(L_{rgb}\), jointly constrain the semantic fusion process of 3D scenes. The external potential field driving ensures semantic consistency across views, while the internal clustering suppresses outlier noise and enhances intra-cluster coherence. Moreover, the internal-clustering term \(L_{cf}\) refines the fine-grained details captured by \(L_{gt}\) and accelerates its convergence. Two hyper-parameters $\lambda_1$ and $\lambda_2$ are manually selected to balance the contribution of external semantic guidance and internal clustering regularization, respectively. Finally, we define the overall loss $L$ as:
\begin{align*}
    L = L_{rgb}+ \lambda_1 L_{gt}+\lambda_2 L_{cf}
\end{align*}
Under the NDFD mechanism, gradient with respect to \(w_k\) not only influence the geometry and appearance of local primitive but also affect the spatial distribution of subsequent gaussian primitives in the backward traversal. The gradient can be obtained by:
\begin{align*}
    \frac{\partial L_{cf}}{\partial w_k} = \sigma_k(W_{k-1}-\mathbf{F}_{k-1}\cdot \mathbf{f}_k) + \sum_{i=k+1}^{N}\sigma_i w_i(1 - \mathbf{f}_i\cdot \mathbf{f}_k)
\end{align*}
The symmetry between the forward and backward passes allows cumulative terms computed during the forward traversal to be directly reused in gradient calculations (see Appendix), eliminating redundant passes and preserving the \(\mathcal{O}(N)\) complexity of both processes.

\label{sec:method}

\section{Experiments}
We implement \textit{FHGS} within a 2DGS-based framework, deploying tailor-made CUDA kernels to accelerate the proposed feature-fusion operations. We use the image embedding of SAM~\cite{sam} as input to the feature. The original 2DGS renderer is retained to export depth-distortion maps, depth maps, normal maps, and mesh reconstructions, which serve as the inputs to our quantitative and qualitative evaluations.  
In the sigmoid activation function, the similarity threshold and slope are empirically fixed to \(\lambda = 0.5\) and \(k = 20\), respectively, ensuring stable binarization of the feature-matching score \(\sigma\) that governs the polarity of gaussian primitive.
For benchmarking, we adopt Feature3DGS~\cite{featuregs} as the baseline. Following its protocol, we report the \(L_1\) feature loss FL1 (lower values indicating higher feature similarity) under the same rendering pipeline, where smaller FL1 values signify better feature fusion. Cross-view consistency is further assessed with the ground-truth entropy metric \(L_{gt}\) (Eq.~\ref{equ:lgt}); lower \(L_{gt}\) scores indicate tighter multi-view alignment.
To ensure fair comparisons, all experiments are executed on a workstation equipped with a single NVIDIA GeForce RTX 4090 (24 GB) and an AMD Ryzen 9 9950X (16 cores). In addition, we use identical feature-extraction pipelines together with the default 2DGS optimizer settings (learning rate, iteration count, batch size) for both the baseline and our method, thereby eliminating performance biases due to hyperparameter tuning or feature-generation differences.

\begin{table}[h]
    \centering
    \caption{Quantitative results comparison on indoor scenes}
    \label{tab:scene_comparison_indoor}
    \resizebox{\textwidth}{!}{%
    \begin{tabular}{@{}lcccccccccccc@{}}
    \toprule
    \multirow{2}{*}{Method} & \multicolumn{4}{c}{DTU-24~\cite{dtu}} & \multicolumn{4}{c}{DTU-37~\cite{dtu}} & \multicolumn{4}{c}{MN360-kitchen~\cite{mipnerf360}}\\
    \cmidrule(lr){2-5} \cmidrule(lr){6-9} \cmidrule(l){10-13}
     & PSNR$\uparrow$& FE$\downarrow$& FL1$\downarrow$& Time$\downarrow$& PSNR$\uparrow$& FE$\downarrow$& FL1$\downarrow$& Time$\downarrow$& PSNR$\uparrow$& FE$\downarrow$& FL1$\downarrow$& Time$\downarrow$\\
    \midrule
    2DGS         & 30.1& 1.35& 0.61& 6.1m& 30.5& 1.31& 0.52& 6.3m& 30.2& 1.32& 0.79& 6.5m\\
    Feature3DGS  & \textbf{31.5}& 0.52& 0.24& 82.2m& \textbf{31.1}& 0.88& 0.31& 73.2m& \textbf{31.7}& 0.63& 0.31& 113.2m\\
    FHGS (\textbf{ours}) & 30.9& \textbf{0.15}& \textbf{0.22}& \textbf{5.2m}& 30.8& \textbf{0.21}& \textbf{0.18} & \textbf{5.7m}& 30.8& \textbf{0.23}& \textbf{0.21} & 5.1m\\
    \bottomrule
    \end{tabular}
    }
\end{table}

\subsection{Comparative experiment}
To verify the generalization and robustness of our method, we conduct systematic experiments on a range of public datasets covering both indoor and outdoor environments. For indoor evaluations, we evaluate our method on DTU (scans 24, 37)~\cite{dtu} and Mip-NeRF 360 (Kitchen)~\cite{mipnerf360}, as the results shown in Table~\ref{tab:scene_comparison_indoor}, while outdoor evaluations are performed on Mip-NeRF 360 (Garden, Stump) and Tanks and Temples (TnT Caterpillar)~\cite{tnt}, the results are shown in the Table~\ref{tab:scene_comparison_outdoor}. All input images are uniformly downsampled to a maximum side length of 1,000 pixels to balance computational efficiency and reconstruction accuracy. Sparse point clouds are initialized with COLMAP~\cite{colmap} are used for SfM, with a fixed iteration count of 10,000 to ensure optimization consistency. During testing, Feature3DGS~\cite{featuregs} failed in TnT~\cite{tnt} due to its huge utilization of GPU memory. 

The experimental results demonstrate that \textit{FHGS} reduces training time by \(15\times\) relative to Feature3DGS, improves performance by \(8\text{–}10\%\) over standard 2DGS, and maintains real-time rendering at \(\ge 60\) FPS. In terms of feature fusion quality, \textit{FHGS} achieves the same performance comparable to Feature3DGS in the FL1 metrics, validating its effectiveness in feature similarity measurement. \textit{FHGS} also exhibits superior FE metrics (lower values denote stronger cross-view consistency), highlighting its advantage in semantic coherence across viewpoints. 

\begin{table}[ht]
    \centering
    \caption{Quantitative results comparison on outdoor scenes}
    \label{tab:scene_comparison_outdoor}
    \resizebox{\textwidth}{!}{%
    \begin{tabular}{@{}lcccccccccccc@{}}
    \toprule
    \multirow{2}{*}{Method} & \multicolumn{4}{c}{COLMAP~\cite{colmap}} & \multicolumn{4}{c}{MN360-Garden~\cite{mipnerf360}} & \multicolumn{4}{c}{TnT-Caterpillar~\cite{tnt}}\\
    \cmidrule(lr){2-5} \cmidrule(lr){6-9} \cmidrule(l){10-13}
     & PSNR$\uparrow$& FE$\downarrow$& FL1$\downarrow$& Time$\downarrow$& PSNR$\uparrow$& FE$\downarrow$& FL1$\downarrow$& Time$\downarrow$& PSNR$\uparrow$& FE$\downarrow$& FL1$\downarrow$& Time$\downarrow$\\
    \midrule
    2DGS         & 27.4& 1.73& 0.83& 10.16m& 31.3& 1.67 & 0.75 & 6.3m & 26.8& 1.72& 0.76& 5.2m\\
    Feature3DGS  & \textbf{28.2}& 0.55& 0.42& 181m& \textbf{31.6}& 0.65 & 0.33 & 155.4m & -& -& -& -\\
    FHGS (\textbf{ours}) & 26.5& \textbf{0.25}& \textbf{0.24}& \textbf{7.8m}& 30.6& \textbf{0.25} & \textbf{0.18} & \textbf{6.1m} & \textbf{26.6}& \textbf{0.21}& \textbf{0.41}& \textbf{5.2m}\\
    \bottomrule
    \end{tabular}
    }
\end{table}

\begin{figure}[htbp]
  \centering
  \includegraphics[width=1.0\textwidth]{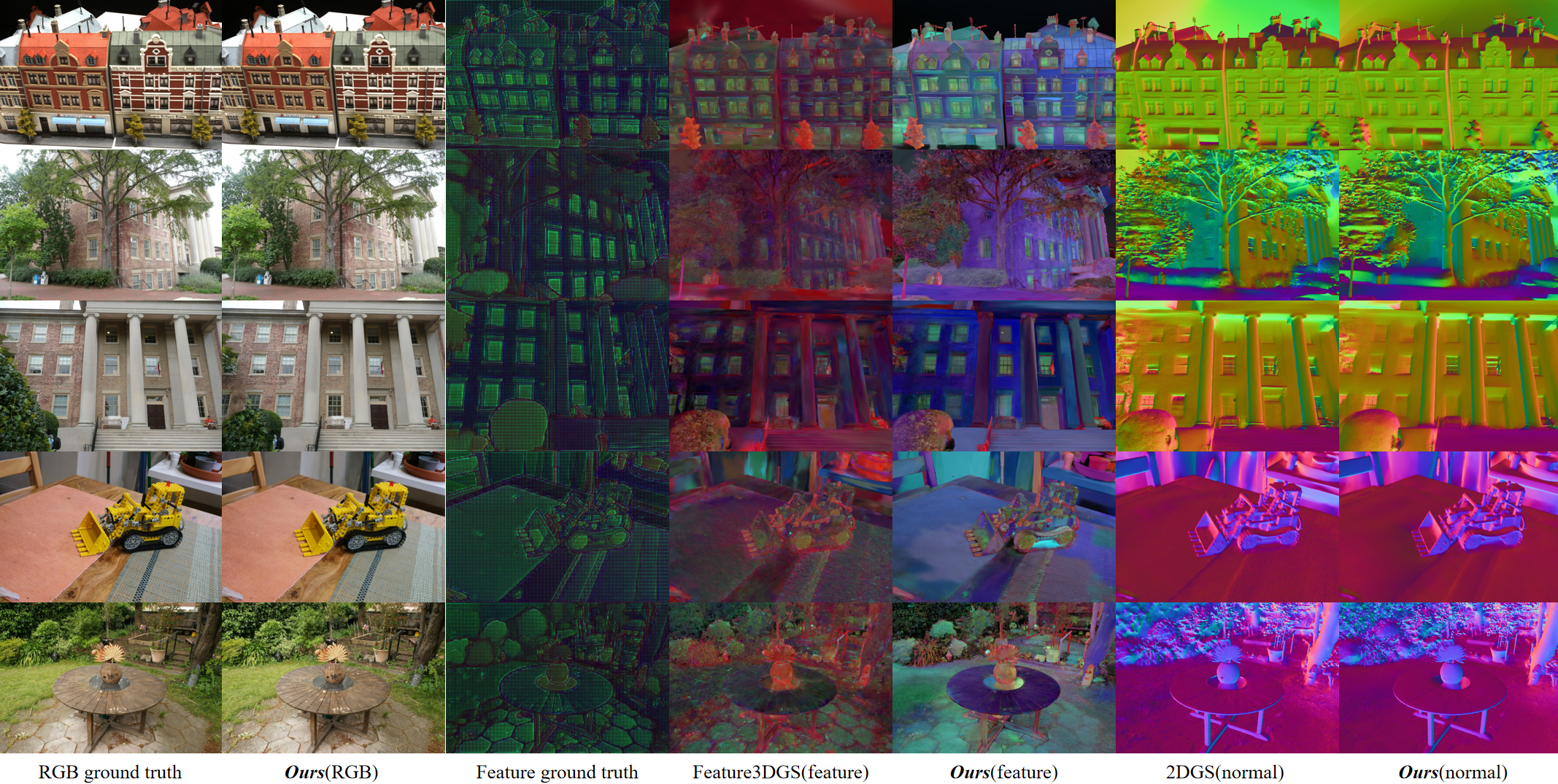} 
  \caption{Qualitative results to compare our \textit{FHGS} with Feature3DGS~\cite{featuregs} in feature map and 2DGS~\cite{2dgs} in normal map.}
  \label{fig:output}
\end{figure}

To visually assess fusion quality, we map channels 15, 28 and 31 of the image embedding features to RGB for rendering. The visualization comparison results in Fig.~\ref{fig:output} further demonstrates the superiority of our \textit{FHGS} which produces uniform feature distributions with minimal noise, smooth semantic transitions, and clear boundaries. For geometric reconstruction, our method effectively suppresses noise and drives geometric structures to converge toward thin planar surfaces, ultimately achieving high-precision surface reconstruction comparable to MVS (see Appendix). 

Our training is faster and uses less GPU memory in Table~\ref{tab:accuracy_completion}. These results conclusively demonstrate that \textit{FHGS} significantly enhances semantic-geometric consistency in 3D scene representations while preserving real-time rendering efficiency through the proposed novel fusion mechanism and optimization strategy. More results of experiments can be found in the Appendix.
\begin{table}[ht]
    \centering
    \caption{Quantitative results between \textit{FHGS}, 3DGS, 2DGS and Feature3DGS on the DTU~\cite{dtu}, we report chamfer distance, PSNR (training-set view), reconstruction time, model size and point number.}
    \label{tab:accuracy_completion}
    \begin{tabular}{@{}lccc cc@{}}
    \toprule
    Methods & CD$\uparrow$& PSNR$\uparrow$& Time$\downarrow$& PN$\downarrow$& MB (Storage)\\
    \midrule
    3DGS         & 1.96 & \textbf{35.76} & 11.2m & 532k & 113 \\
    2DGS         & 0.83 & 33.42 & 5.5m & 342k  & \textbf{52} \\
    Feature3DGS  & 1.85 & 35.25 & >24h & 642k & 745 \\
    FHGS (\textbf{ours})  & \textbf{0.75} & 34.21 & \textbf{4.8m}  & \textbf{196k} & 183\\
    \bottomrule
    \end{tabular}
\end{table}

\subsection{Ablation Study}

\begin{figure}[htbp]
  \centering
  \includegraphics[width=1\textwidth]{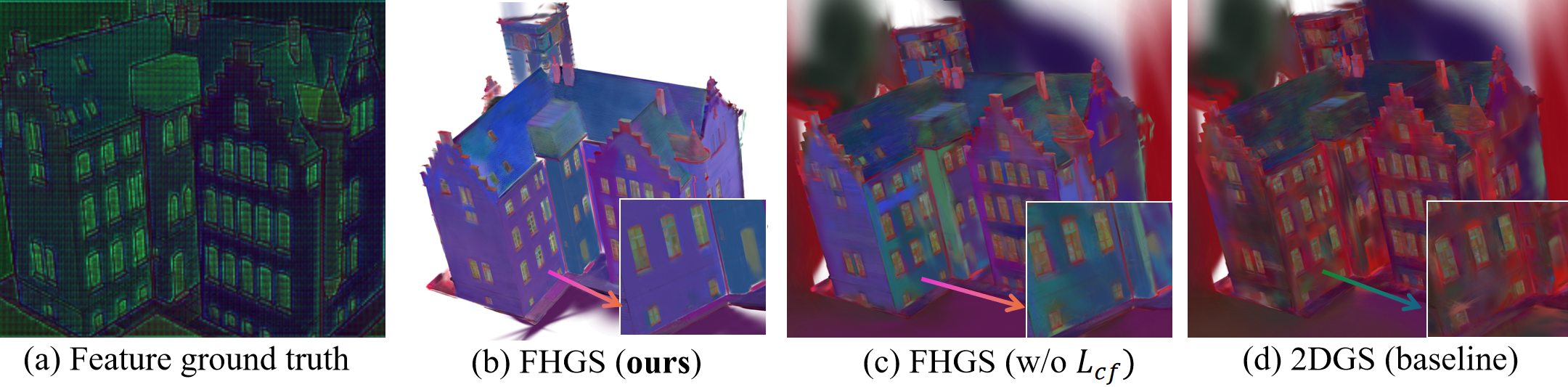} 
  \caption{Ablation study on the scan24 of DTU dataset~\cite{dtu}.}
  \label{fig:Ablation}
\end{figure}

The ablation study is conducted on the scan24 of DTU dataset~\cite{dtu} with 10,000 training iterations to investigate the effects of the loss functions \(L_{gt}\) and \(L_{cf}\) on feature fusion, geometric reconstruction, and optimization efficiency (illustrated in Table~\ref{tab:ablation_study}). Fig.~\ref{fig:Ablation} (a) illustrates the result of image embedding from SAM~\cite{sam}. The experimental results indicate that these two loss terms serve complementary roles:  As illustrated in the Fig.~\ref{fig:Ablation} (d), when \(L_{gt}\) and \(L_{cf}\) are both disabled, and visualizing the feature through the default rendering logic, the resulting feature map diverges markedly from the ground truth and appears cluttered. When only \(L_{cf}\) is removed, although the optimization proceeds faster, the model suffers from semantic contamination and overfitting at the surface level. As shown in the Fig.~\ref{fig:Ablation} (c), numerous valid gaussian primitives are incorrectly discarded, leading to excessive transparency in the reconstructed geometry and severe degradation in reconstruction quality. When both  \(L_{gt}\) and \(L_{cf}\) are jointly applied, the framework achieves an optimal balance: the feature consistency metric FE improves, geometric structures converge toward thin and planar forms, and convergence speed increases. Fig.~\ref{fig:Ablation} (b) has shown that under this configuration, the distribution of the feature field is uniform and dense, semantic boundaries are sharp and well-defined, and the reconstructed surfaces retain detailed geometric information. These findings validate the effectiveness of the dual-loss collaborative optimization strategy. 
\vspace{-1.5em}

\begin{table}[h]
    \centering
    \caption{Quantitative analysis of ablation experiments on DTU scan24}
    \label{tab:ablation_study}
    \begin{tabular}{@{}lccccc@{}}
    \toprule
    Methods & PSNR$\uparrow$ & FE$\downarrow$ & FL1$\downarrow$ & Time$\downarrow$ & PN$\downarrow$\\
    \midrule
    FHGS (\textbf{ours})& \textbf{30.9}& 0.15& \textbf{0.16}& 5.2m & \textbf{214k}\\
    FHGS (w/o $L_{cf}$)& 27.2& \textbf{0.10}& 0.21& \textbf{4.3m}& 217k\\
    2DGS (baseline)& 30.1& 1.35 & 0.46 & 6.1m & 329k\\
    \bottomrule
    \end{tabular}
\end{table}
\label{sec:exp}
\section{Conclusion and Discussion}

We introduce a Gaussian splatting based framework named \textit{FHGS}, which incorporates a non‑differentiable feature‑driven regularization term to enforce multi‑view semantic consistency. \textit{FHGS} markedly boosts multi‑view feature alignment and geometric reconstruction quality while maintaining real‑time performance, as demonstrated by extensive experiments on diverse indoor and outdoor datasets. While our \textit{FHGS} successfully achieves multi‑view consistent and accurate geometric reconstruction, it still has some limitations: our methods remains sensitive to the manually tuned similarity‑activation parameters $lambda$ and $k$; its hash–table and cumulative‑weight buffers incur considerable GPU memory in large‑scale scenes. In future work, we plan to explore adaptive parameter learning strategies to reduce dependence on manual tuning, and to develop memory-efficient and compact representations to enhance scalability in large-scale environments.

\label{sec:conclusion}

\begin{ack}
This study was supported by the InnoHK initiative of the Innovation and Technology Commission of the Hong Kong Special Administrative Region Government via the Hong Kong Centre for Logistics Robotics, and in part by the Research Grants Council of Hong Kong SAR under Grants 14206821, 14217922 and 14209623.
\end{ack}


\newpage
\appendix

\section{Detailed Explanation of Non-Differentiable Feature Driving (NDFD)}\label{sec:appendix_NDNF}

\begin{algorithm}[H]
\caption{General Feature Fusion and Densification Framework}
\begin{algorithmic}          
\algrenewcommand\algorithmiccomment[1]{\hfill$\triangleright$~#1}

\State $\mathbf{F}_{gt} \gets \textsc{GetFeatureFromSAM}(\mathbf{I}_{gt})$ \Comment{Features}
\State $(p,\mathcal{H}) \gets \textsc{GetPointFromSFM}(\mathbf{I}_{gt})$  \Comment{Positions \& Hash}
\State $\mathbf{f} \gets \textsc{PointsFeatureFusion}(p,\mathbf{F},\mathcal{H})$ \Comment{Point Features}
\State $(\mathbf{R},\mathbf{S},c,op) \gets \textsc{InitAttributes}()$      \Comment{Rotations, Scales, Color, Opacity}
\State $i \gets 0$                                                         \Comment{Iteration Counter}

\While{not converged}
    \State $(m,\mathbf{I}_{gt}^m,\mathbf{F}_{gt}^m) \gets \textsc{SampleTrainingView}()$ \Comment{Camera, Image, Feature}
    \State $\mathbf{I}_{re}^m \gets \textsc{drf}(p,\mathbf{R},\mathbf{S},op,c,m)$         \Comment{Rasterization}
    \State $L_{rgb} \gets \textsc{loss}(\mathbf{I}_{gt}^m,\mathbf{I}_{re}^m)$             \Comment{Photometric Loss}
    \State $(L_{gt},L_{cf}) \gets \textsc{ndfd}(p,\mathbf{R},\mathbf{S},op,\mathbf{f},\mathbf{F}_{gt}^m,m)$ \Comment{Semantic Loss}
    \State $L \gets L_{rgb}+L_{gt}+L_{cf}$                                                \Comment{Total Loss}
    \State $(p,\mathbf{R},\mathbf{S},op,c) \gets \textsc{Adam}(\nabla L)$                 \Comment{Update}

    \If{\textsc{IsRefinementIteration}(i)}
        \State \textsc{Densification}$(p,\mathbf{R},\mathbf{S},op,c,\mathbf{f})$            \Comment{Adaptive Density}
    \EndIf
    \State $i \gets i+1$
\EndWhile
\end{algorithmic}
\end{algorithm}

\emph{Details of the Rasterizater}: Our implementation builds directly on the GPU rasterizer proposed in 3D Gaussian Splatting. Following that design, the image plane of size $w\!\times\!h$ is partitioned into $16\times16$ \,px tiles. Each gaussian primitive that overlaps a tile is \emph{duplicated} for that tile and assigned a 64‑bit key whose lower 32 bits encode depth and upper bits encode the tile index. A single parallel radix sort on these keys resolves global depth order and produces a compact, per‑tile, depth‑sorted list of instances; a second pass identifies the start–end range for each tile (see \textsc{CullGaussian}, \textsc{DuplicateWithKeys}, and \textsc{SortByKeys} in 3DGS). This eliminates sequential primitive traversal and maximizes GPU utilization.

\begin{algorithm}[H]
\caption{Non-Differentiable Feature Driving Mechanism}
\begin{algorithmic}           
\Function{NDFD}{$p,\mathbf{R},\mathbf{S},op,\mathbf{f},\mathbf{F}_{gt}^m,m$}
    \State $\mathbf{x}\gets\textsc{Homogenization}(m)$                     \Comment{Camera Homogenization}
    \State $g\gets\mathcal{G}_i(\mathbf{u}(\mathbf{x}))\gets\textsc{2DScreenGaussians}(p,\mathbf{R},\mathbf{S},\mathbf{x})$ \Comment{Screen‐space Gaussians}
    \State $T\gets\textsc{CreateTiles}(m)$                                 \Comment{Tile Grid}
    \State $(I,K)\gets\textsc{DuplicateWithKeys}(g,T)$                     \Comment{Indices \& Keys}
    \State \textsc{SortByKeys}$(K,I)$                                       \Comment{Global Sort}
    \State $T_r\gets\textsc{IdentifyTileRanges}(T,K)$                      \Comment{Tile Ranges}
    \State $L_{gt}\gets0,\;L_{cf}\gets0$                                   \Comment{Initilize Loss Buffers}

    \ForAll{$t\in T$}
        \ForAll{$i\in t$}
            \State $r\gets\textsc{GetTileRange}(T_r,t)$                    \Comment{Index Range in $K$}
            \ForAll{$j\in r$}
                \State $\sigma_j\gets\textsc{Sigmoid}(\mathbf{F}_{gt}^m(i),f_j)$ \Comment{Polarity Response}
                \State $w_j\gets\textsc{WeightCalc}(g_j,op_j)$                   \Comment{Opacity‐weighted Area}
                \State $L_{gt}[i]\mathrel{+}= \textsc{epfc}(w_j,f_j,\sigma_j)$   \Comment{External Potential}
                \State $L_{cf}[i]\mathrel{+}= \textsc{icd}(w_j,f_j,\sigma_j)$    \Comment{Internal Clustering}
            \EndFor
        \EndFor
    \EndFor

    \State \Return $(L_{gt},L_{cf})$
\EndFunction
\end{algorithmic}
\label{alg:ndfd}
\end{algorithm}

\emph{Non-Differentiable Feature Driving (NDNF)}: Alg.~\ref{alg:ndfd} augments the aforementioned rasterizer with a feature‑centric branch that runs entirely on the sorted gaussian primitives lists and never invokes $\alpha$ blending. Given the current view~$m$, camera homogenization first projects gaussian primitives means into screen space, after which key generation and radix sorting produce per‑tile ranges. For every pixel $i$ in a tile $t$, we then traverse the corresponding range~$r$ in front‑to‑back order. A sigmoid activation $\sigma_j=\textsc{Sigmoid}(\mathbf{F}_{gt}^m(i),f_j)$ converts the cosine similarity between the frozen feature $f_j$ of the $j‑th$ gaussian primitives and the ground‑truth embedding $\mathbf{F}_{gt}^m(i)$ into a charge‑like polarity. The raster weight $w_j$, which combines projected area and opacity exactly as in $\alpha$ blending, is accumulated only by this feature branch. Then, two loss terms are computed: the external‑potential loss $L_{gt}$ attracts $\sigma_j$‑weighted features toward $\mathbf{F}_{gt}^m(i)$, whereas the internal‑clustering loss $L_{cf}$ applies the cumulative‑feature rule to penalize incoherent neighbors. These loss buffers are initialized once per frame and updated atomically in the innermost loop, so no intermediate feature image is rendered. During back‑propagation, gradients propagate solely through the weights $w_j$, which reuse the same cumulative prefix employed for $\alpha$‑blending in the forward traversal, thereby retaining the $\mathcal{O}(N)$ complexity of the original rasterizer.

\subsection{Derivation of the feature similarity}\label{sec:appendix_formula_lcf}

\emph{Internal clustering loss $L_{cf}$}: For a given pixel \(p\), let \(\{(w_i,f_i)\}_{i=1}^{N}\) be the set of gaussian primitive whose screen‑space footprints cover that pixel, where \(w_i\) is the weight and \(f_i\in\mathbb{R}^{d}\) is the frozen semantic feature of the \(i-th\) primitive.  
The internal‑clustering loss:
\begin{equation}
 L_{cf} = \sum_{i=1}^{N} \sum_{j=1}^{N} w_i \, w_j \, \left(1 - \cos \langle f_i \cdot f_j \rangle \right) 
\end{equation}
computes the entropy of the local feature distribution by accumulating the weighted cosine dissimilarity between every pair of primitives. 
The process of minimizing \(L_{cf}\)  pushes feature vectors of neighboring primitive to align, suppresses noisy outliers, and tightens semantic coherence within the pixel neighborhood. And simultaneously, this process allows primitives belonging to different objects to repel each other through their low cosine similarity.

We further convert it to an \(\mathcal{O}(N)\) backward traversal by noting that feature vectors are normalized, therefore
\(\cos\langle f_i,f_j\rangle = f_i\!\cdot\!f_j\). We can rearranged the representation of $L_{cf}$ as:
\begin{equation}
    \begin{split}
        L_{cf}
        & = \sum_{i=1}^{N} \sum_{j=1}^{i-1} \sigma_i \, w_i \, w_j \left(1 - \cos \langle f_i \cdot f_j \rangle \right) \\
        &= \sum_{i=1}^{N} \sigma_i \, w_i \left( \sum_{j=1}^{i-1} w_j - \sum_{j=1}^{i-1} w_j \, f_j \cdot f_i \right)     \\
        &= \sum_{i=1}^{N} \sigma_i w_i \left( \sum_{j=1}^{i-1} w_j - \sum_{j=1}^{i-1} w_j f_j \cdot f_i \right)
        \\
        &= \sum_{i=1}^{N} \sigma_i \, w_i \left( W_{i-1} - F_{i-1} \cdot f_i \right)
    \end{split}
\end{equation}
where the cumulative weight \(W_{i-1}\) and cumulative feature \(F_{i-1}\) are updated one time in each step during the front‑to‑back blend. The final form of $L_{cf}$ retains the physical meaning of pairwise semantic attraction–repulsion so evaluates in \(\mathcal{O}(N)\) time. The cumulative values of $W_{N-1}$ and $F_{N-1}$ are recorded.

\subsection{Calculation process of the gradients}\label{sec:appendix_formula_grad}

As derived in the main text, we obtain the partial derivatives:
\begin{equation}
    \frac{\partial L_{cf}}{\partial w_k} 
    = \sigma_k \left( W_{k-1} - \mathbf{F}_{k-1} \cdot \mathbf{f}_k \right)
    + \sum_{i=k+1}^{N} \sigma_i w_i \left(1 - \mathbf{f}_i \cdot \mathbf{f}_k \right)
\end{equation}
In the forward pass the gaussian primitive is processed in descending depth order from the farthest to the nearest with respect to the camera. The backward pass visits the same primitive in the reverse order.  
Because the index \(k\) is defined with respect to the forward ordering, we re‑index the backward traversal by a new counter \(q=1,\dots ,N\). Exploiting this forward–backward symmetry, the gradient of the internal‑clustering loss with respect to the weight of the current primitive can be rewritten as:
\begin{equation}
    \frac{\partial L_{cf}}{\partial w_q} 
    =  \sigma_q\big( (W_{N}-W_{q}) - (\mathbf{F}_{N} - \mathbf{F}_{q})\cdot \mathbf{f}_q \big)
    + \sum_{i=1}^{q-1} \sigma_i w_i \big(1 - \mathbf{f}_i \cdot \mathbf{f}_q \big)
\end{equation}
Based on the ${w}_i = \alpha_i T_i$,  $\alpha_i = op_i{\mathcal{G}}_i(\mathbf{u}(\mathbf{x}))$,  $T_i = \prod_{j=1}^{i-1}(1-\alpha_j)$, we can obtain:
\begin{equation}
    \frac{\partial L_{gt}}{\partial \alpha_k} = \frac{\partial L_{gt}}{\partial w_k} \cdot \frac{\partial w_k}{\partial \alpha_k} = T_k \cdot \frac{\partial L_{gt}}{\partial w_k} - \frac{1}{1 - \alpha_k} \sum_{i=k+1}^{N} \frac{\partial L_{gt}}{\partial w_i} \cdot w_i
\end{equation}
Analogously to the equation above, we can also obtain:
\begin{equation}
    \frac{\partial L_{gt}}{\partial \alpha_q} = \frac{\partial L_{gt}}{\partial w_q} \cdot \frac{\partial w_q}{\partial \alpha_q} = T_q \cdot \frac{\partial L_{gt}}{\partial w_q} - \frac{1}{1 - \alpha_q} \sum_{i=0}^{q-1} \frac{\partial L_{gt}}{\partial w_i} \cdot w_i
\end{equation}

\section{Comprehensive Results of Experiments}\label{sec:comprehensive_results}

\begin{figure}[htbp]
  \centering
  \includegraphics[width=1.0\textwidth]{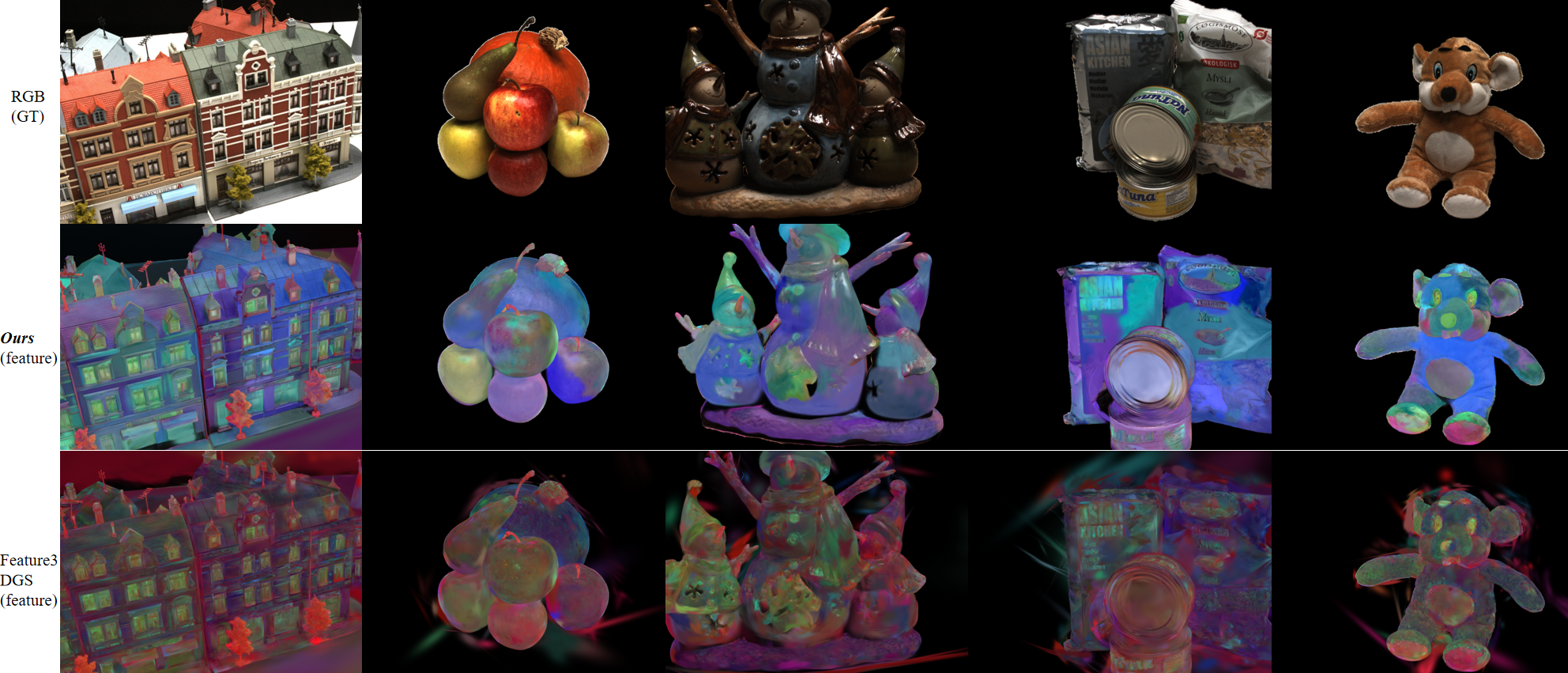} 
  \caption{Qualitative results to compare our \textit{FHGS} with Feature3DGS in feature field. The results shown that \textit{FHGS} achieves better feature extraction with more uniform feature distributions, shaper boundaries and cleaner background.}
  \label{fig:app_1}
\end{figure}

We conduct a detailed comparison between our method and Feature3DGS on the DTU indoor dataset. As shown in the Fig.~\ref{fig:app_1}, our method yields more uniform feature distributions and sharper boundaries. Moreover, it effectively suppresses background clutter, which remains prominent in Feature3DGS. The enhanced clarity and selectivity of our features also benefit downstream tasks such as segmentation and reconstruction. These observations highlight the strength of our feature driving mechanism in promoting structural coherence and semantic focus.

\begin{figure}[h]
  \centering
  \begin{subfigure}[b]{\textwidth}
    \centering
    \includegraphics[width=1.0\textwidth]{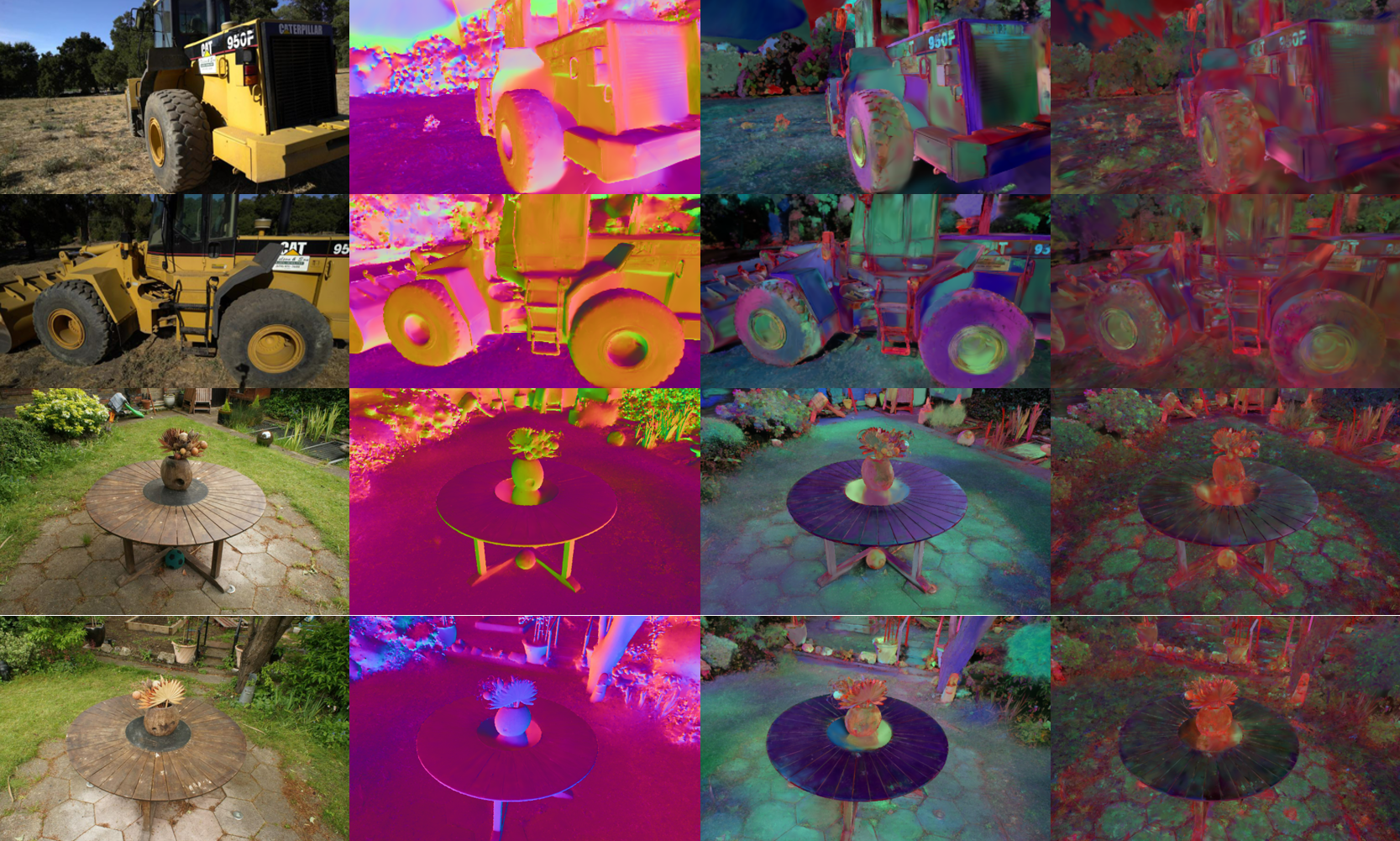}
  \end{subfigure}
  
  \begin{subfigure}[b]{\textwidth}
    \centering
    \includegraphics[width=1.0\textwidth]{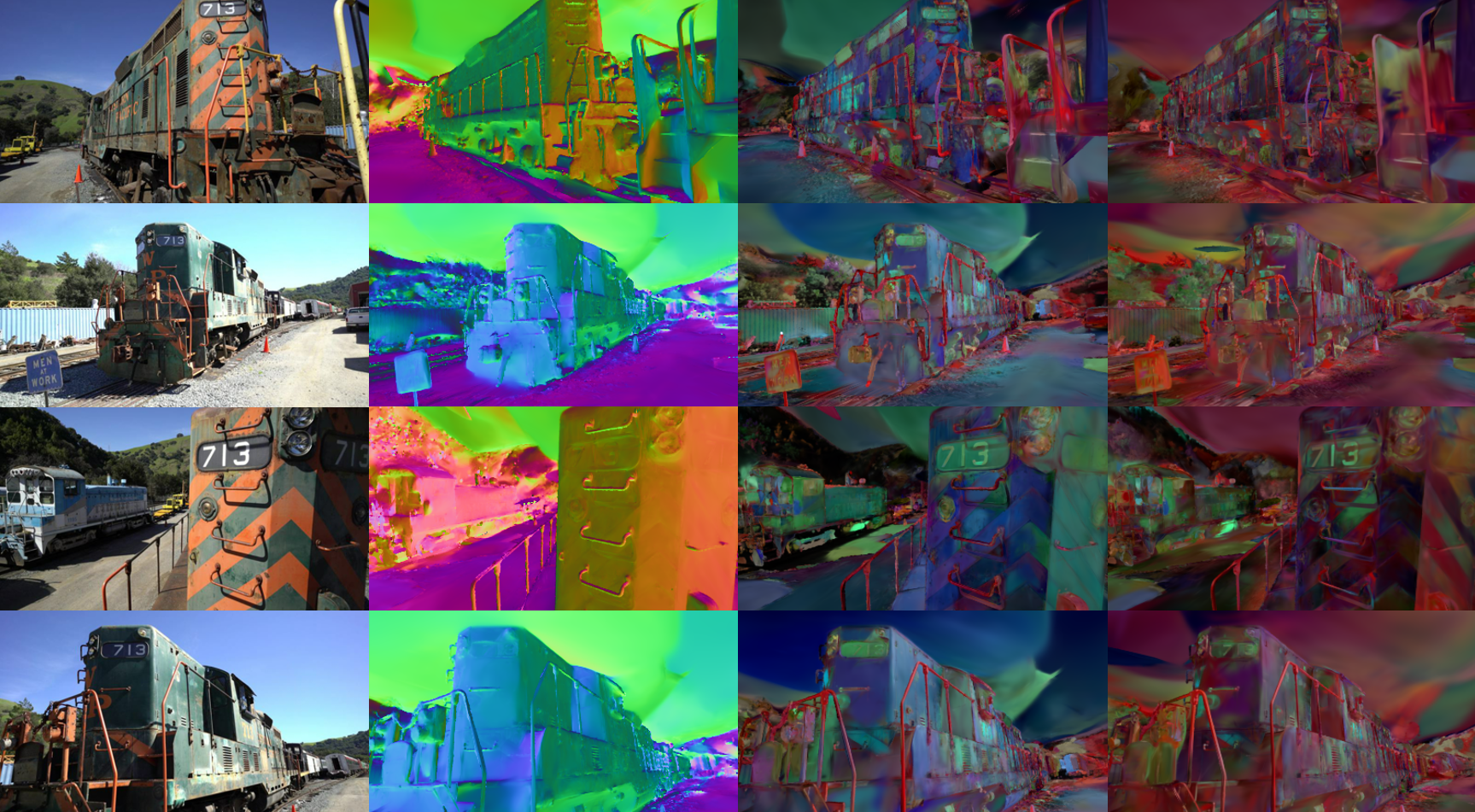}
  \end{subfigure}
  
  \begin{subfigure}[b]{\textwidth}
    \centering
    \includegraphics[width=1.0\textwidth]{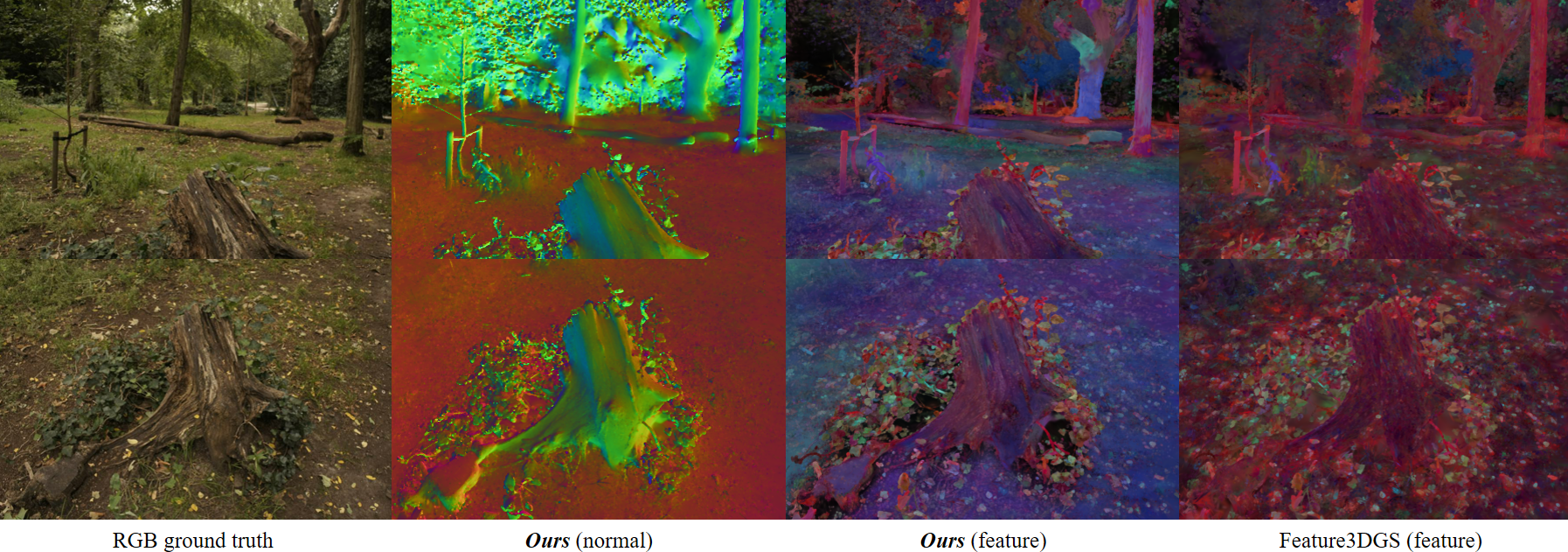}
  \end{subfigure}
  
  \caption{Qualitative comparison on outdoor scenes from TnT and MipNeRF360}
  \label{fig:outdoor_combined}
\end{figure}

We further evaluate our method against Feature3DGS on challenging outdoor scenes from the TnT and MipNeRF360 datasets. 
As shown in Fig.~\ref{fig:outdoor_combined}, our method consistently delivers more coherent and spatially uniform feature fields, with significantly clearer object boundaries and effective suppression of background noise. In addition to semantic features, we also visualize the surface normal maps extracted from our reconstruction, which exhibit plausible geometric structures and fine-grained surface details. These results demonstrate the robustness of our method under natural lighting, large-scale geometry, and high-frequency textures, confirming its generalization to diverse outdoor environments.

\end{document}